\documentclass{article}

\usepackage{PRIMEarxiv}

\usepackage[utf8]{inputenc} % allow utf-8 input
\usepackage[T1]{fontenc}    % use 8-bit T1 fonts
\usepackage{hyperref}       % hyperlinks
\usepackage{url}            % simple URL typesetting
\usepackage{booktabs}       % professional-quality tables
\usepackage{amsfonts}       % blackboard math symbols
\usepackage{nicefrac}       % compact symbols for 1/2, etc.
\usepackage{microtype}      % microtypography  
\usepackage{lipsum}
\usepackage{fancyhdr}       % header
\usepackage{graphicx}       % graphics
\graphicspath{{media/}}     % organize your images and other figures under media/ folder

\usepackage{orcidlink, natbib, amsmath, amsfonts, algorithm, algpseudocode, multirow}
\newcommand{\code}[1]{\texttt{#1}}

\newcommand{\Prob}{\mathsf{P}}

\title{NUBO: A Transparent Python Package for Bayesian Optimisation
\thanks{Journal reference: 
Diessner, M., Wilson, K. J., \& Whalley, R. D. (2025). NUBO: A Transparent Python Package for Bayesian Optimization. Journal of Statistical Software, 114(1), 1–28. https://doi.org/10.18637/jss.v114.i01}
}

\author{
  Mike Diessner~\orcidlink{0000-0001-9838-0862}, Kevin J. Wilson~\orcidlink{0000-0001-9450-862X} \\
  Newcastle University \\
  Newcastle upon Tyne \\
   \And
  Richard D. Whalley~\orcidlink{0000-0001-6388-5895} \\
  Queen’s University Belfast \\
  Belfast \\
}

\begin{document}
\maketitle

\begin{abstract}
NUBO, short for Newcastle University Bayesian Optimisation, is a Bayesian 
optimisation framework for the optimisation of expensive-to-evaluate black-box 
functions, such as physical experiments and computer simulators. Bayesian 
optimisation is a cost-efficient optimisation strategy that uses surrogate 
modelling via Gaussian processes to represent an objective function and 
acquisition functions to guide the selection of candidate points to approximate 
the global optimum of the objective function. NUBO itself focuses on 
transparency and user experience to make Bayesian optimisation easily accessible 
to researchers from all disciplines. Clean and understandable code, precise 
references, and thorough documentation ensure transparency, while user experience 
is ensured by a modular and flexible design, easy-to-write syntax, and careful 
selection of Bayesian optimisation algorithms. NUBO allows users to tailor 
Bayesian optimisation to their specific problem by writing the optimisation loop 
themselves using the provided building blocks. It supports sequential single-point, 
parallel multi-point, and asynchronous optimisation of bounded, constrained, 
and/or mixed (discrete and continuous) parameter input spaces. Only algorithms 
and methods that are extensively tested and validated to perform well are included 
in NUBO. This ensures that the package remains compact and does not 
overwhelm the user with an unnecessarily large number of options. The package is 
written in Python but does not require expert knowledge of 
Python to optimise your simulators and experiments. NUBO is 
distributed as open-source software under the BSD 3-Clause licence.
\end{abstract}

% keywords can be removed
\keywords{Bayesian optimisation \and black-box optimisation \and surrogate model \and
  Gaussian process \and Monte Carlo \and design of experiments \and Python}

\section{Introduction} \label{sec:intro}

The optimisation of expensive black-box functions is a common problem encountered 
by researchers in a wide range of disciplines, such as engineering, computing, 
and natural sciences. These functions are characterised by an unknown or not 
analytically solvable mathematical expression and high costs of evaluation. The 
principal way to gather information about a black-box function is to provide it 
with some inputs and observe its corresponding output. However, this process 
produces high costs, for example, material costs for physical experiments, 
computing costs for simulators, or time costs in general \citep{Frazier2018}. 
Many optimisation algorithms, such as Adam \citep{Kingma2014}, L-BFGS-B 
\citep{Zhu1997}, and differential evolution \citep{Storn1997}, rely either on 
derivative information about the objective function or large numbers of function 
evaluations. Neither is typically feasible when working with an expensive 
black-box function, requiring us to search elsewhere for a cost-effective and 
sample-efficient alternative.

Bayesian optimisation takes a surrogate model-based approach with the aim of 
optimising expensive black-box functions in a minimum number of function 
evaluations. Although the genesis of Bayesian optimisation can be traced back to 
the middle of the 20th century \citep{Kushner1964, Zilinskas1975, Mockus1975, 
  Mockus1989}, it gained considerable popularity in the last two decades 
\citep{Jones1998, Snoek2012, Shahriari2015, Frazier2018}. In recent years, it 
has been applied to simulators and experiments in various research areas. For 
example, Bayesian optimisation was used in the field of computational fluid 
dynamics to maximise the drag reduction via the active control of blowing 
actuators \citep{Diessner2022, OConnor2023, Mahfoze2019}, in chemical engineering 
for molecular design, drug discovery, molecular modelling, electrolyte design, 
and additive manufacturing \citep{Wang2022}, and in computer science to fine-tune 
hyper-parameters of machine learning models \citep{Wu2019} and for architecture 
search of neural networks \citep{White2021}.

With NUBO, we provide an open-source implementation of Bayesian optimisation 
aimed at researchers with expertise in disciplines other than statistics and 
computer science. To ensure that our target audience can understand and use 
Bayesian optimisation to its full potential, NUBO focuses particularly on 
(a)~transparency through clean and understandable code, precise references, and 
thorough documentation and (b)~user experience through a modular and flexible 
design, easy syntax, and a careful selection of implemented algorithms. Various 
Python packages for Bayesian optimisation exist as listed in 
Table~\ref{tab:overview}. Most of them only support sequential single-point 
optimisation, i.e.,~every point suggested by the algorithm has to be evaluated by 
the objective function before moving on to the next iteration. However, in many 
cases, parallelism can be exploited to speed up the optimisation process. For 
example, consider a simulator that can be run in parallel. Evaluating all points 
in parallel would save time as it would only take as long as evaluating a single 
point sequentially. pyGPGO \citep{Jimenez2017}, 
bayes\_opt\footnote{The package is also known under the name 
bayesian-optimization.} \citep{Nogueira2014}, Spearmint 
\citep{Spearmint2023}, and SMAC3 \citep{Lindauer2022} do not allow parallel 
multi-point optimisation. Furthermore, Spearmint is not modular, resulting 
in less flexible implementations and giving the user less control when tailoring 
Bayesian optimisation to unique research problems. To our knowledge, the closest 
available package to NUBO is BoTorch \citep{Balandat2020} as it also 
supports parallel and asynchronous optimisation through the use of Monte Carlo 
approximations of the acquisition functions. However, compared to the lightweight 
implementation of NUBO, BoTorch uses a very large code base that makes 
code comprehension difficult, as it often requires retracing various functions 
and objects through a large number of files. This can be quantified by the huge 
codebase represented in Table~\ref{tab:overview} as the total number of lines of 
code\footnote{The total number of lines of code does not include comments, blank 
lines and files that are irrelevant to the actual algorithms, such as examples, 
tests and test functions.}: NUBO implements Bayesian optimisation in only 
1,322 lines of code over 20 files, while BoTorch uses 38,419 lines of 
code---roughly 29 times more than NUBO---and spreads them between 160 files. 
It also provides a large number of functions and methods that enforce decisions 
non-expert users do not have the knowledge and experience to make. NUBO 
lightens this burden of the user by limiting itself to the most important methods. 
Table~\ref{tab:overview} also includes GPyOpt \citep{gpyopt2016}; however, 
it is no longer maintained and has recently been archived.

The number of code lines regards the underlying code bases of the packages and 
not the lines of code a user must write to apply Bayesian optimisation. When 
talking about a transparent implementation, the former is a better proxy as it 
reflects the complexity of the whole package, that is, all functions and 
algorithms of the package. If a package has many thousands of lines---such as 
BoTorch---it is intuitive that it is more complex and thus harder to fully 
comprehend than a package with only a few hundred lines of code---such as 
NUBO. The number of code lines it takes to apply Bayesian optimisation is 
less informative as it can easily be biased and distorted. Consider, for example, 
a very complex algorithm with many lines of code. It would be possible to wrap 
this algorithm into one function that can be called with one line of code. While 
this reduces the lines of code, it does not change the algorithm's complexity. 
Thus, the comparison in this article focuses on the number of lines of the 
underlying code bases to give an idea of the size and complexity of the packages. 

\begin{table}[t!]
    \centering
    \begin{tabular}{@{}lrrrrrr@{}}
        \hline
Type & Modular & Sequential & Parallel & Asynchronous & Lines of code & Version \\
        \hline
NUBO      & Yes     & Yes     & Yes      & Yes    &  1,322     & 1.0.3 \\ 
BoTorch   & Yes     & Yes     & Yes      & Yes    & 38,419     & 0.8.4 \\ 
bayes\_opt& Yes     & Yes     & No       & No     &  1,241     & 1.4.3 \\
SMAC3     & Yes     & Yes     & No       & No     & 11,217     & 2.0.0 \\
pyGPGO    & Yes     & Yes     & No       & No     &  2,029     & 0.5.0 \\
GPyOpt    & Yes     & Yes     & Yes      & No     &  4,605     & 1.2.6 \\
Spearmint & No      & Yes     & No       & No     &  3,662     &   0.1 \\
        \hline
    \end{tabular}
\caption{\label{tab:overview} Overview of available Bayesian optimisation packages 
in Python. We compare whether individual packages have a modular design 
and support sequential single-point, parallel multi-point, and asynchronous 
optimisation. We also list the number of lines of code of the core package 
(without comments, examples, tests, etc.) and the version number.}
\end{table}

\begin{figure}[t!] 
\label{fig:packages}
\centering
\includegraphics[width=0.9\linewidth]{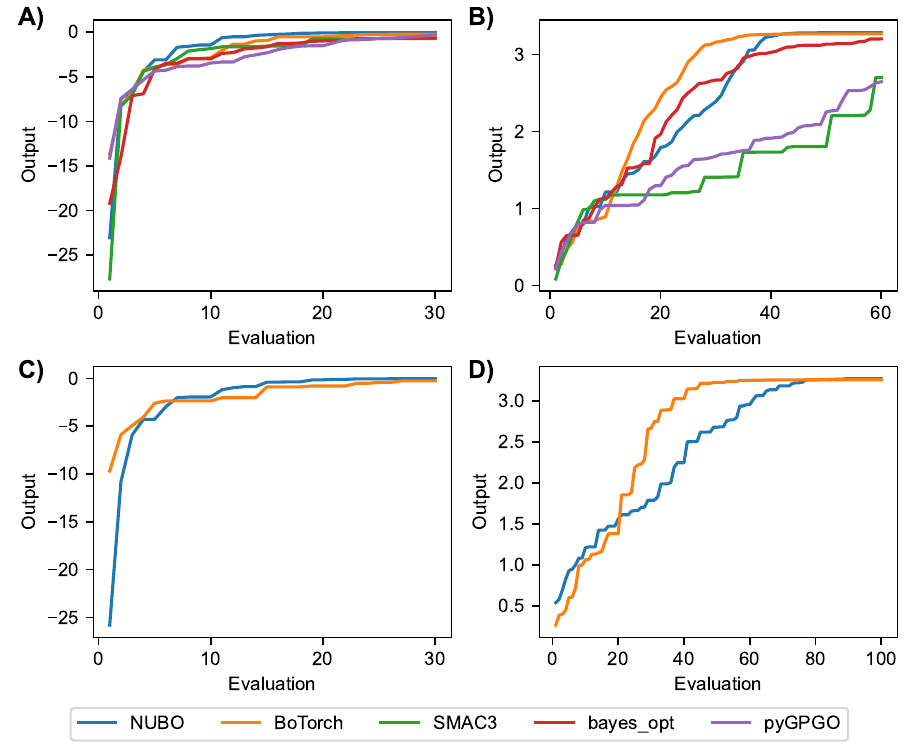}
\caption{Comparison of different Python packages for Bayesian 
optimisation. A)~Sequential single-point optimisation on the 2D Levy function;  
B)~Sequential single-point optimisation on the 6D Hartmann function; C)~Parallel 
multi-point optimisation with a batch size of four on the 2D Levy function;  
D)~Parallel multi-point optimisation with a batch size of four on the 6D Hartmann 
function.}
\end{figure}

Although it is difficult to provide an exhaustive comparison of the relative 
efficiency of each of the packages, we have undertaken a limited 
comparison\footnote{All comparisons were run on an Apple Mac mini with a M2 chip 
and 16 GB memory.} of the following form. We compare NUBO to four of the 
packages mentioned above---BoTorch, bayes\_opt, SMAC3 and 
pyGPGO---representing a reasonably wide range of complexity. All methods 
use Gaussian processes (introduced in Section~\ref{sec:bo-gp}) as the surrogate 
model and upper confidence bound (introduced in Section~\ref{sec:bo-acq}) as the 
acquisition function. Please see the replication materials published alongside 
this article for further details on the algorithms and the benchmarking. Two 
synthetic test functions from \cite{Surjanovic} were selected to benchmark the 
five packages. The first row of plots in Figure~\ref{fig:packages} compares the 
performance of sequential single-point optimisation on A)~the two-dimensional 
Levy function and B)~the six-dimensional Hartmann function. The second row of 
plots compares the performance of parallel multi-point optimisation with batches 
of four points on C)~the two-dimensional Levy function and D)~the six-dimensional 
Hartmann function. All functions are negated to transform them from their initial 
minimisation problem into a maximisation problem in line with the convention of 
Bayesian optimisation. The plots provide the best observation at the current 
evaluation, averaging over ten replication runs. The results show that all 
packages converge towards the global optimum of 0.00 for the Levy and 3.32 for 
the Hartmann function. While NUBO requires more evaluations to find the 
global optimum for the Hartmann function (B) and D)) than the more complex 
packages such as BoTorch, it gets closest to the true optimum in all cases 
after all evaluations and shows low variance in these results 
(Table~\ref{tab:pkg-means}). These results show that the simplicity of 
NUBO's implementation does not come at a cost in performance. NUBO 
can outperform packages with a similar level of complexity, such as pyGPGO 
and bayes\_opt, and compares well against more complex packages, such as 
BoTorch and SMAC3. This is not to say that NUBO is the superior 
package for any problem, but rather that NUBO performs competitively while 
focusing on a transparent and simple design. This makes NUBO a good 
candidate for the optimisation of expensive black-box functions in the 
sciences---such as physical experiments and computer simulators---where 
transparency is vital.

\begin{table}[t!]
    \centering
    \begin{tabular}{@{}lrrrr@{}}
\toprule
\multirow{3}{*}{} & \multicolumn{2}{c}{Sequential} & \multicolumn{2}{c}{Parallel}\\
\cmidrule{2-5}
                  & 2D Levy           & 6D Hartmann      & 2D Levy           
                & 6D Hartmann \\
\midrule
NUBO        & -0.04 ($\pm$0.06) & 3.28 ($\pm$0.06) & -0.04 ($\pm$0.04) 
                & 3.27 ($\pm$0.06) \\ 
BoTorch     & -0.21 ($\pm$0.20) & 3.27 ($\pm$0.07) & -0.27 ($\pm$0.21) 
                & 3.26 ($\pm$0.06) \\ 
SMAC3       & -0.71 ($\pm$0.58) & 2.70 ($\pm$0.38) & -                & - \\ 
bayes\_opt  & -0.64 ($\pm$0.74) & 3.20 ($\pm$0.13) & -                & - \\ 
pyGPGO      & -0.28 ($\pm$0.31) & 2.64 ($\pm$1.05) & -                & - \\ 
\bottomrule
    \end{tabular}
\caption{\label{tab:pkg-means} Comparison of different Python packages 
for Bayesian optimisation. The best observations averaged across the ten runs 
with corresponding standard errors are given for each package.}
\end{table}

However, the time NUBO requires to complete one iteration with a maximum of 
2.20s for D)~is, on average, higher than for the other packages 
(Table~\ref{tab:pkg-times}). While this might be important for some areas of 
optimisation, it is negligible when it comes to the optimisation of expensive 
black-box functions, as these functions are much more resource-intensive to 
evaluate. Thus, the small number of additional seconds that NUBO requires 
per iteration is insignificant compared to the resources required to conduct an 
experiment or a simulation.

\begin{table}[t!]
    \centering
    \begin{tabular}{@{}lrrrr@{}}
        \toprule
\multirow{3}{*}{}&\multicolumn{2}{c}{Sequential}&\multicolumn{2}{c}{Parallel}\\
        \cmidrule{2-5}
                          & 2D Levy & 6D Hartmann & 2D Levy & 6D Hartmann \\
        \hline
        NUBO        & 0.60s   & 1.88s       & 0.07s   & 2.20s \\ 
        BoTorch     & 0.09s   & 0.22s       & 0.00s   & 0.19s \\ 
        SMAC3       & 0.08s   & 0.25s       & -       & - \\ 
        bayes\_opt  & 0.14s   & 0.24s       & -       & - \\ 
        pyGPGO      & 0.23s   & 0.65s       & -       & - \\  
        \bottomrule
    \end{tabular}
\caption{\label{tab:pkg-times} Comparison of different Python packages 
for Bayesian optimisation. The elapsed time per iteration averaged across the 
ten runs is given for each package.}
\end{table}

Besides implementations in Python, there are also some implementations 
in other programming languages. For example, rBayesianOptimization 
\citep{Yan2021} and ParBayesianOptimization \citep{Wilson2022} implement 
basic Bayesian optimisation algorithms for hyper-parameter tuning similar to 
bayes\_opt and pyGPGO in R. ParBayesianOptimization 
provides additional support for parallel optimisation and follows 
\cite{Wilson2018}. 

The remainder of this paper is structured as follows. In Section~\ref{sec:bo}, we 
introduce the Bayesian optimisation algorithm, including Gaussian processes that 
form the surrogate model and acquisition functions that guide the optimisation. 
The implementation of Bayesian optimisation in NUBO is discussed in 
Section~\ref{sec:nubo} before we illustrate how NUBO can be used to optimise 
expensive black-box functions through a non-trivial case study in 
Section~\ref{sec:case-study}. Finally, we draw conclusions and give an outlook on 
future work in Section~\ref{sec:conclusion}.

%% -- Manuscript ---------------------------------------------------------------

\section{Bayesian optimisation} \label{sec:bo}

Bayesian optimisation aims to solve the $d$-dimensional maximisation problem
\begin{equation} \label{eq:max-problem}
    \boldsymbol x^* = \arg \max_{\boldsymbol x \in \mathcal{X}} f(\boldsymbol x),
\end{equation}
where the input space is usually continuous and bounded by a hyper-rectangle 
$\mathcal{X} \in [a, b]^d$ with $a, b \in \mathbb{R}$. The function 
$f(\boldsymbol x)$ is most commonly a derivative-free, expensive-to-evaluate 
black-box function that allows inputs $\boldsymbol x_i$ to be queried and outputs 
$y_i$ to be observed without gaining any further insights into the underlying 
system \citep{Frazier2018}. We assume any noise $\epsilon$ introduced when taking 
measurements to be independent and identically distributed Gaussian noise 
$\epsilon \sim \mathcal{N} (0, \sigma^2)$ such that $y_i = f(\boldsymbol x_i) + 
  \epsilon$. Hence, a set of $n$ pairs of input data points and corresponding 
observations is defined as
\begin{equation*} \label{eq:data}
    \mathcal{D}_n = \{(\boldsymbol x_i, y_i)\}_{i=1}^n 
\end{equation*}
and we further define training inputs as the matrix 
$\boldsymbol X_n = \{\boldsymbol x_i \}_{i=1}^n$ and their training outputs as 
the vector $\boldsymbol y_n = \{y_i\}_{i=1}^n$. Simulators and experiments in 
various disciplines can be formulated to fit this description, including but not 
limited to the examples given in the introduction.

Bayesian optimisation \citep{Frazier2018, Gramacy2020, Jones1998, Shahriari2015, 
  Snoek2012} is a surrogate model-based optimisation algorithm that aims to 
maximise the objective function $f(\boldsymbol x)$ in a minimum number of function 
evaluations. Typically, the objective function does not have a known or analytical 
solvable mathematical expression, and every function evaluation is expensive. Such 
problems require a cost-effective and sample-efficient optimisation strategy. 
Bayesian optimisation meets these criteria by representing the objective function 
through a surrogate model $\mathcal{M}$, often a Gaussian process. This 
representation can be used to find the input points to be evaluated sequentially 
by maximising a criterion specified through an acquisition function 
$\alpha (\cdot)$. A popular criterion is the upper confidence bound (UCB). This 
acquisition function can be classed as an optimistic acquisition function that 
considers the upper bound of the uncertainty around the surrogate model's 
prediction to be true \citep{Shahriari2015}. Bayesian optimisation is performed 
in a loop, where training data is used to fit the surrogate model before the next 
point suggested by the acquisition function is evaluated and added to the training 
data (see the Algorithm~\ref{alg:bo} below). The process then restarts and gathers 
more information about the objective function with each iteration. Bayesian 
optimisation is run for as many iterations as the evaluation budget $N$ allows, 
as shown in Algorithm~\ref{alg:bo}, until a satisfactory solution is found or 
until a predefined stopping criterion is met.

\begin{algorithm}[t!]
\caption{Bayesian optimisation algorithm}\label{alg:bo}
\begin{algorithmic}

\Require Evaluation budget $N$, number of initial points $n_0$, surrogate model 
  $\mathcal{M}$, acquisition function $\alpha$.

\State Sample $n_0$ initial training data points $\boldsymbol X_0$ via a space-
  filling design \citep{McKay2000} and gather observations $\boldsymbol y_0$.
\State Set $n = 0$.

\State Set $\mathcal{D}_n = \{ \boldsymbol X_0, \boldsymbol y_0 \}$.

\While{$n \leq N -n_0$}

\State Fit surrogate model $\mathcal{M}$ to training data $\mathcal{D}_n$.  
\State Find $\boldsymbol x_n^*$ that maximises an acquisition criterion $\alpha$ 
  based on model $\mathcal{M}$.  
\State Evaluate $\boldsymbol x_n^*$ observing $y_n^*$ and add to $\mathcal{D}_n$.  
\State Increment $n$.

\EndWhile

\State \Return Point $\boldsymbol x^*$ with highest observation $y^*$.
\end{algorithmic}
\end{algorithm}

\begin{figure}[t!] 
\label{fig:bo}
\centering
\includegraphics[width=0.5\linewidth]{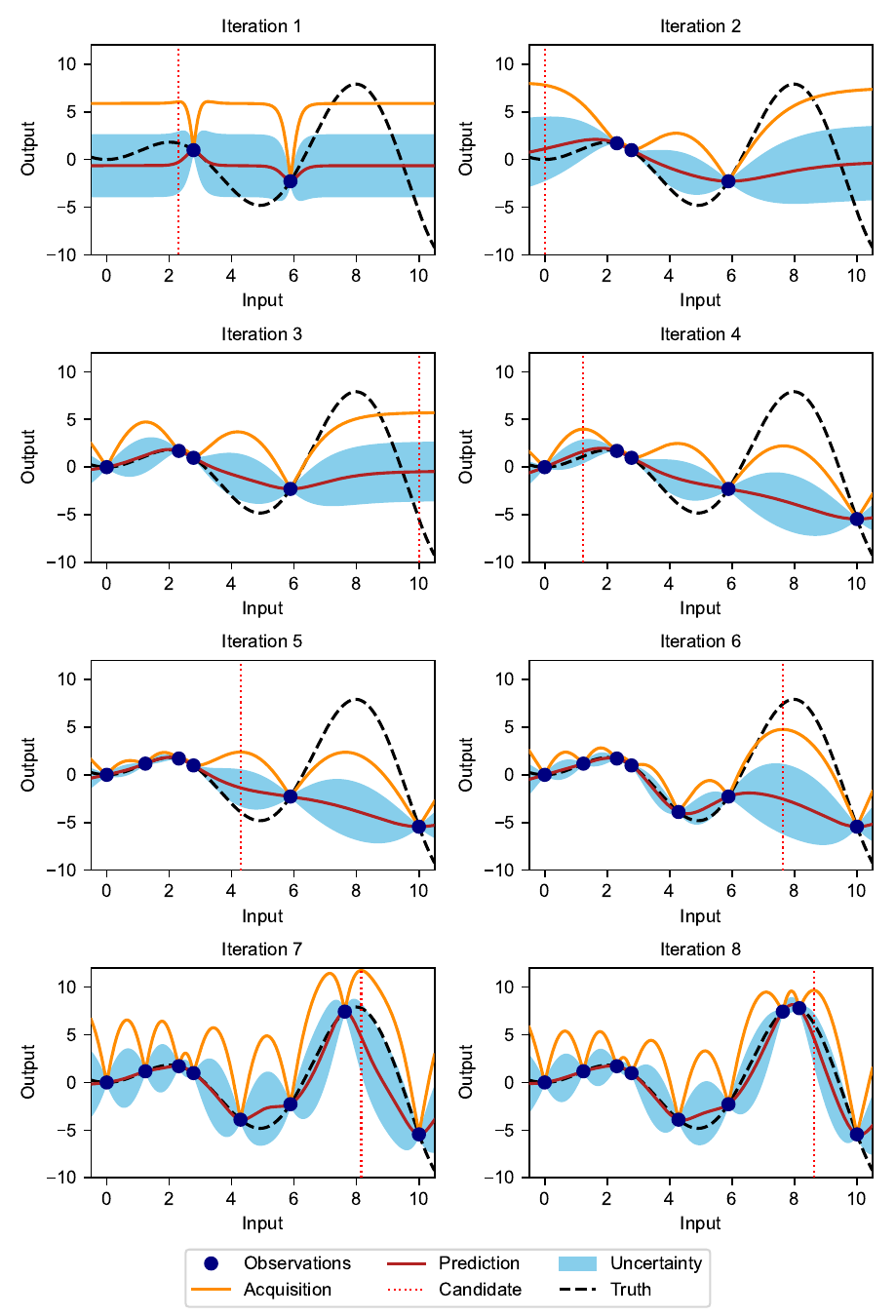}
\caption{Bayesian optimisation applied to a 1-dimensional function with one local 
and one global maximum. Upper confidence bound is used as the acquisition 
function. The input space is bounded by $[0, 10]$.}
\end{figure}

Figure~\ref{fig:bo} illustrates how the Bayesian optimisation algorithm works for 
an optimisation loop that runs for eight iterations and starts with two initial 
training points. In this example, NUBO finds an approximation of the global 
optimum ($x = 8$) for a simple 1-dimensional problem on iteration seven. The 
surrogate model uses the available observations to provide a prediction and 
associated uncertainty (here shown as 95\% confidence intervals around the 
prediction). This is our best estimate of the underlying objective function. This 
estimate is then used in the acquisition function to evaluate which input value 
is likely to return a high output. Maximising the acquisition function provides 
the next candidate point to be observed from the objective function before it is 
added to the training data and the whole process is repeated. Figure~\ref{fig:bo} 
shows how the surrogate model converges to the true objective function with each 
iteration. The acquisition function covers the input space by exploring regions 
with high uncertainty and exploiting regions with a high prediction. This property, 
known as the exploration-exploitation trade-off, is a cornerstone of the 
acquisition functions provided in NUBO.

\subsection{Gaussian processes} \label{sec:bo-gp}

A popular choice for the surrogate model $\mathcal{M}$ that acts as a 
representation of the objective function $f(\boldsymbol x)$ is a Gaussian process 
\citep{Gramacy2020, Rasmussen2006}, a flexible non-parametric regression model. 
A Gaussian process is a finite collection of random variables that has a joint 
Gaussian distribution and is defined by a prior mean function 
$\mu_0(\boldsymbol x) : \mathcal{X} \mapsto \mathbb{R}$ and a prior covariance 
kernel $\Sigma_0(\boldsymbol x, \boldsymbol x') : \mathcal{X} \times \mathcal{X} 
  \mapsto \mathbb{R}$ resulting in the prior distribution
\begin{equation*} \label{eq:gp-prior}
  f(\boldsymbol X_n) \sim \mathcal{N} (m(\boldsymbol X_n), K(\boldsymbol X_n, 
  \boldsymbol X_n)),
\end{equation*}
where $m(\boldsymbol X_n) = \mu_0(\boldsymbol X_n)$ is the mean vector of length 
$n$ over all training inputs and $K(\boldsymbol X_n, \boldsymbol X_n) = \Sigma_0 
  (\boldsymbol X_n, \boldsymbol X_n)$ is the $n \times n$ covariance matrix 
between all training inputs.

Popular choices for the prior mean function $\mu_0(\cdot)$ are the zero and 
constant mean functions given in Equations~\ref{eq:zero-mean} and 
\ref{eq:const-mean}. For the prior covariance function $\Sigma_0(\cdot, \cdot)$, 
the squared exponential kernel, also called the radial basis function (RBF) 
kernel, and the Mat\'ern $\frac{5}{2}$ kernel are popular options. The covariance 
kernels in Equations~\ref{eq:rbf} and \ref{eq:matern} are based on the distance 
$r = \rvert \boldsymbol x - \boldsymbol x' \lvert$ and have two parameters: the 
signal variance $\sigma^2_f$, sometimes also referred to as the output-scale, and 
the characteristic length-scale $l$. The former will scale the function with 
larger values resulting in a larger deviation from its mean, while the latter 
indicates for how long function values are correlated along the input axes, the 
smaller the length-scale $l$ the shorter the correlation \citep{Gramacy2020, 
  Rasmussen2006}.
\begin{equation} \label{eq:zero-mean}
    \mu_{\text{zero}}(\boldsymbol x) = 0 
\end{equation}
\begin{equation} \label{eq:const-mean}
    \mu_{\text{constant}}(\boldsymbol x) = c
\end{equation}
\begin{equation} \label{eq:rbf}
    \Sigma_{\text{RBF}} (\boldsymbol x, \boldsymbol x') = \sigma^2_f \exp 
      \left( -\frac{r^2}{2l^2} \right)
\end{equation}
\begin{equation} \label{eq:matern}
    \Sigma_{\text{Mat\'ern}} (\boldsymbol x, \boldsymbol x') = \sigma^2_f 
    \left( 1 + \frac{\sqrt{5}r}{l} + \frac{5r^2}{3l^2} \right) \exp 
    \left( -\frac{\sqrt{5}r}{l} \right)
\end{equation}
Covariance functions can be extended to include one characteristic length-scale 
$l_d$ for each input dimension $d$. In this case, input dimensions with large 
length-scales are correlated for longer distances and are less relevant for 
changes in the prediction. This means that varying the values of the input 
dimension affects the prediction little. Input dimensions with small 
length-scales are correlated for shorter distances, and even small changes in the 
input values can affect the prediction significantly. Gaussian processes with 
covariance functions that include multiple length-scales are characterised by 
automatic relevance determination (ARD) of the input dimensions \citep{Neal1996}. 
Here, the inverse of the length-scales can be interpreted as the relevance of the 
corresponding dimensions \citep{Rasmussen2006}. The Gaussian process will estimate 
large length-scales for irrelevant dimensions, automatically assigning them less 
importance. 

The posterior distribution for $n_*$ test points $\boldsymbol X_*$ can be 
computed as the multivariate Gaussian distribution conditional on training data 
$\mathcal{D}_n$
\begin{equation*} \label{eq:gp-posterior}
    f(\boldsymbol X_*) \mid \mathcal{D}_n, \boldsymbol X_* \sim \mathcal{N} 
    \left(\mu_n (\boldsymbol X_*), \sigma^2_n (\boldsymbol X_*) \right)
\end{equation*}
\begin{equation*} \label{eq:gp-posterior-mean}
    \mu_n (\boldsymbol X_*) = K(\boldsymbol X_*, \boldsymbol X_n) 
    \left[ K(\boldsymbol X_n, \boldsymbol X_n) + \sigma^2_y I \right]^{-1} 
    (\boldsymbol y - m (\boldsymbol X_n)) + m (\boldsymbol X_*)
\end{equation*}
\begin{equation*} \label{eq:gp-posterior-cov}
    \sigma^2_n (\boldsymbol X_*) = K (\boldsymbol X_*, \boldsymbol X_*) - 
    K(\boldsymbol X_*, \boldsymbol X_n) \left[ K(\boldsymbol X_n, \boldsymbol X_n) 
    + \sigma^2_y I \right]^{-1} K(\boldsymbol X_n, \boldsymbol X_*),
\end{equation*}
where $m(\boldsymbol X_*)$ is the mean vector of length $n_*$ over all test 
inputs, $K(\boldsymbol X_*, \boldsymbol X_n)$ is the $n_* \times n$  covariance 
matrix, $K(\boldsymbol X_n, \boldsymbol X_*)$ is the $n \times n_*$ covariance 
matrix, $K(\boldsymbol X_*, \boldsymbol X_*)$ is the $n_* \times n_*$ covariance 
matrix between training inputs $\boldsymbol X_n$ and test inputs $\boldsymbol X_*$ 
respectively, and $\sigma^2_y$ is the noise variance of the Gaussian process.

The hyper-parameters $\theta$ of the Gaussian process, for example, the constant 
$c$ in the mean function, the signal variance $\sigma^2_f$ and characteristic 
length-scales $\boldsymbol l$ in the covariance kernel, and the noise variance 
$\sigma^2_y$, can be estimated by maximising the log-marginal likelihood in 
Equation~\ref{eq:gp-ll} via maximum likelihood estimation (MLE) 
\citep{Rasmussen2006}.
\begin{multline} \label{eq:gp-ll}
    \log \Prob(\boldsymbol y_n \mid \boldsymbol X_n) = -\frac{1}{2} 
    (\boldsymbol y_n - m(\boldsymbol X_n))^\top [K(\boldsymbol X_n, 
    \boldsymbol X_n) + \sigma^2_y I]^{-1} (\boldsymbol y_n - m(\boldsymbol X_n))\\
    - \frac{1}{2} \log \lvert K(\boldsymbol X_n, \boldsymbol X_n) + 
    \sigma^2_y I \rvert - \frac{n}{2} \log 2 \pi
\end{multline}

\subsection{Acquisition functions} \label{sec:bo-acq}

Acquisition functions use the posterior distribution of the Gaussian process to 
compute a criterion that assesses if a test point is a good potential candidate 
point to evaluate via the objective function $f(\boldsymbol x)$. Thus, maximising 
the acquisition function suggests the test point that, based on the current 
training data $\mathcal{D}_n$, has the highest potential and information gain to 
get closer to the global optimum while exploring the input space. To do this, an 
acquisition function $\alpha (\cdot)$ balances exploration and exploitation. The 
former is characterised by areas with no or only a few observed data points where 
the uncertainty of the Gaussian process is high, and the latter by areas where 
the posterior mean of the Gaussian process is large. This exploration-exploitation 
trade-off ensures that Bayesian optimisation does not converge to the first 
(potentially local) maximum it encounters but efficiently explores the full input 
space.

\subsubsection{Analytical acquisition functions}

NUBO supports two of the most popular acquisition functions whose 
performance has been demonstrated in both theoretical and empirical research. 
Expected improvement (EI) \citep{Jones1998} selects points with the biggest 
potential to improve on the current best observation, while upper confidence 
bound (UCB) \citep{Srinivas2010} takes an optimistic view of the posterior 
uncertainty and assumes it to be true to a user-defined level. Expected 
improvement (EI) is defined as
\begin{equation} \label{eq:acq-ei}
    \alpha_{\text{EI}} (\boldsymbol X_*) = \left(\mu_n(\boldsymbol X_*) - 
    y^{best} \right) \Phi(z) + \sigma_n(\boldsymbol X_*) \phi(z),
\end{equation}
where $z = \frac{\mu_n(\boldsymbol X_*) - y^{best}}{\sigma_n(\boldsymbol X_*)}$, 
$\mu_n(\cdot)$ and $\sigma_n(\cdot)$ are the mean and the standard deviation of 
the posterior distribution of the Gaussian process, $y^{best}$ is the current 
best observation, and $\Phi (\cdot)$ and $\phi  (\cdot)$ are the cumulative 
distribution function and probability density function of the standard normal 
distribution $\mathcal{N}(0, 1)$.

The upper confidence bound (UCB) acquisition function can be computed as
\begin{equation} \label{eq:acq-ucb}
    \alpha_{\text{UCB}} (\boldsymbol X_*) = \mu_n(\boldsymbol X_*) + \sqrt{\beta} 
    \sigma_n(\boldsymbol X_*),
\end{equation}
where $\beta$ is a predefined trade-off parameter, and $\mu_n(\cdot)$ and 
$\sigma_n(\cdot)$ are the mean and the standard deviation of the posterior 
distribution of the Gaussian process. For guidance on the choice of $\beta$, 
consult the theoretical properties in \cite{Srinivas2010} or empirical 
conclusions in \cite{Diessner2022}.

Both of these acquisition functions can be maximised with a deterministic 
optimiser, such as L-BFGS-B \citep{Zhu1997} for bounded unconstrained problems or 
SLSQP \citep{Kraft1994} for bounded constrained problems. However, the use of 
analytical acquisition functions is restricted to sequential single-point problems 
for which every point suggested by Bayesian optimisation is observed via the 
objective function $f( \boldsymbol x)$ immediately before the optimisation loop 
is repeated.

\subsubsection{Monte Carlo acquisition functions}

For parallel multi-point batches or asynchronous optimisation, the analytical 
acquisition functions are, in general, intractable. To use Bayesian optimisation 
in these cases, NUBO supports the approximation of the analytical 
acquisition function through Monte Carlo sampling \citep{Snoek2012, Wilson2018}.

The idea is to draw a large number of samples directly from the posterior 
distribution and then approximate the acquisition functions by averaging these 
Monte Carlo samples. This method is made viable by the reparameterisation of the 
acquisition functions and the computation of samples from the posterior 
distribution via base samples randomly drawn from a standard normal distribution 
$z \sim \mathcal{N} (0, 1)$. Thus, the analytical acquisition functions from in 
Equations~\ref{eq:acq-ei} and \ref{eq:acq-ucb} can be approximated as
\begin{equation*} \label{eq:acq-mc-ei}
    \alpha_{\text{EI}}^{\text{MC}} (\boldsymbol X_*) = \max 
    \left(ReLU(\mu_n(\boldsymbol X_*) + \boldsymbol L \boldsymbol z - y^{best}) 
    \right)
\end{equation*}
\begin{equation*} \label{eq:acq-mc-ucb}
    \alpha_{\text{UCB}}^{\text{MC}} (\boldsymbol X_*) = \max 
    \left(\mu_n(\boldsymbol X_*) + \sqrt{\frac{\beta \pi}{2}} \lvert 
    \boldsymbol L \boldsymbol z \rvert \right),
\end{equation*}
where $\mu_n(\cdot)$ is the mean of the posterior distribution of the Gaussian 
process, $\boldsymbol L$ is the lower triangular matrix of the Cholesky 
decomposition of the covariance matrix $\boldsymbol L \boldsymbol L^\top = 
  K(\boldsymbol X_n, \boldsymbol X_n)$, $\boldsymbol z$ are samples from the 
multivariate standard normal distribution $\mathcal{N} (\boldsymbol 0, 
  \boldsymbol I)$, $y^{best}$ is the current best observation, $\beta$ is the 
  user-defined trade-off parameter, and $ReLU (\cdot)$ is the rectified linear 
  unit function that zeros all values below zero and leaves the rest unchanged.

Due to the randomness in the Monte Carlo samples, these acquisition functions can 
only be optimised by stochastic optimisers, such as Adam \citep{Kingma2014}. 
However, there is some empirical evidence that fixing the base samples for 
individual Bayesian optimisation loops does not affect the performance negatively 
\citep{Balandat2020}. This method would allow deterministic optimisers, such as 
L-BFGS-B \citep{Zhu1997} and SLSQP \citep{Kraft1994}, to be used but could 
potentially introduce bias due to sampling randomness.

Two optimisation strategies for multi-point batches are proposed in the literature 
\citep{Wilson2018}: The first is a joint optimisation approach, where the 
acquisition functions are optimised over all points of the batch simultaneously. 
The second option is a greedy sequential approach where one point after another 
is selected, holding all previous points fixed until the batch is full. Empirical 
evidence shows that both methods approximate the acquisition successfully. 
However, the greedy approach seems to have a slight edge over the joint strategy 
for some examples \citep{Wilson2018}. It is also faster to compute for larger 
batches.

Asynchronous optimisation \citep{Snoek2012} leverages the same property as 
sequential greedy optimisation: the pending points that have not yet been 
evaluated can be added to the test points but are treated as fixed. In this way, 
they affect the joint multivariate normal distribution but are not considered 
directly in the optimisation. Asynchronous optimisation is particularly beneficial 
for objective functions for which the evaluation time varies. In these cases, the 
optimisation can be continued while some points are still being evaluated.

%% -- NUBO ---------------------------------------------------------------------

%% - Virtually all JSS manuscripts list source code along with the generated
%%   output. The style files provide dedicated environments for this.
%% - In R, the environments {Sinput} and {Soutput} - as produced by Sweave() or
%%   or knitr using the render_sweave() hook - are used (without the need to
%%   load Sweave.sty).
%% - Equivalently, {CodeInput} and {CodeOutput} can be used.
%% - The code input should use "the usual" command prompt in the respective
%%   software system.
%% - For R code, the prompt "R> " should be used with "+  " as the
%%   continuation prompt.
%% - Comments within the code chunks should be avoided - these should be made
%%   within the regular LaTeX text.

\section[NUBO]{NUBO} \label{sec:nubo}

NUBO is a Bayesian optimisation package in Python that focuses 
on transparency and user experience to make Bayesian optimisation accessible to 
researchers from a wide range of disciplines whose area of expertise is not 
necessarily statistics or computer science. With this overall goal in mind, 
NUBO ensures transparency by implementing clean and comprehensible code, 
precise references and thorough documentation within this research article and on 
our website at \url{www.nubopy.com}. We avoid implementations of overly complex 
and convoluted functions and objects that require the retracing of individual 
elements through multiple files to be fully understood. We prioritise user 
experience defined by a modular and flexible design that can be intuitively 
tailored to unique problems, easy-to-read and write syntax, and a careful 
selection of Bayesian optimisation algorithms. The latter is important as we try 
not to overwhelm the user with a larger number of options but rather focus on 
what is essential to optimise computer simulators and physical experiments 
successfully.

To create a powerful package with good longevity, it is important to start with 
a strong foundation. NUBO is built upon the Torch\footnote{See 
\url{https://pytorch.org/} for documentation and 
\url{https://pypi.org/project/torch/} for package information on PyPI.} ecosystem 
\citep{Paszke2019} that provides a strong scientific computation framework for 
working with tensors, a selection of powerful optimisation algorithms, such as 
\code{torch.Adam} \citep{Kingma2014}, automatic differentiation capabilities to 
compute gradients of acquisition functions via \code{torch.autograd}, and GPU 
acceleration. Furthermore, GPyTorch\footnote{See \url{https://gpytorch.ai/} 
for documentation and \url{https://pypi.org/project/gpytorch/} for package 
information on PyPI.} \citep{Gardner2018}, the package we use to implement 
Gaussian processes for our surrogate modelling, is also based in Torch and 
combines seamlessly with NUBO. We borrow the L-BFGS-B \citep{Zhu1997} and 
SLSQP \citep{Kraft1994} optimisation algorithms from SciPy\footnote{See 
\url{https://scipy.org/} for documentation and 
\url{https://pypi.org/project/scipy/} for package information on PyPI.} 
\citep{scipy} for the deterministic optimisation of the acquisition functions and 
use NumPy\footnote{See \url{https://numpy.org/} for documentation and 
\url{https://pypi.org/project/numpy/} for package information on PyPI.} 
\citep{Harris2020} to make data suitable for these optimisers.

NUBO and all its required dependencies can be installed from the 
Python Package Index (PyPI) \citep{pypi} with the packet installer 
pip \citep{pip} via the terminal:
\begin{verbatim}
    pip install nubopy
\end{verbatim}

\subsection{Gaussian processes} \label{sec:nubo-gp}

NUBO uses the GPyTorch \citep{Gardner2018} package to implement 
Gaussian processes for surrogate modelling. While GPyTorch allows the 
definition of many different Gaussian processes through its various mean 
functions, covariance kernels, and methods for hyper-parameter estimation, we 
provide a predefined Gaussian process in the \code{nubo.models} module that 
follows the work of \cite{Snoek2012}. The \code{GaussianProcess} is specified by 
a constant mean function and the Mat\'ern $\frac{5}{2}$ ARD kernel that, due to 
its flexibility, is well suited for practical optimisation as it can represent a 
wide variety of real-world objective functions. The code below implements a 
Gaussian process and estimates its hyper-parameters from some training inputs 
\code{x\_train} and training outputs \code{y\_train} by maximising the log-marginal 
likelihood in Equation~\ref{eq:gp-ll} with the \code{fit\_gp} function. The 
hyper-parameters include the constant in the mean function, the output-scale and 
length-scales in the covariance kernel, and the noise in the Gaussian likelihood. 
The training inputs and training outputs are specified as a \code{torch.Tensor} 
of size $n \times d$ and length $n$, respectively, where $n$ is the number of 
points and $d$ is the number of input dimensions. Calling the function 
\code{fit\_gp} results in a trained Gaussian process that can subsequently be 
used for Bayesian optimisation.
\begin{verbatim}
    >>> from nubo.models import GaussianProcess, fit_gp
    >>> from gpytorch.likelihoods import GaussianLikelihood
    >>> 
    >>> 
    >>> likelihood = GaussianLikelihood()
    >>> gp = GaussianProcess(x_train, y_train, likelihood = likelihood)
    >>> fit_gp(x_train, y_train, gp = gp, likelihood = likelihood)
\end{verbatim}
While Gaussian processes are capable of estimating noise, for example, 
observational noise occurring when taking measurements from the data, we might 
prefer specifying the noise explicitly if it is known. In these cases, we can 
exchange the \code{GaussianLikelihood} for the \code{FixedNoiseGaussianLikelihood} 
and specify the noise for each training point. The 
\code{FixedNoiseGaussianLikelihood} allows us to decide if any additional noise 
should be estimated by setting the \code{learn\_additional\_noise} attribute to 
\code{True} or \code{False}. The snippet below fixes the observational noise of 
each training point at 2.5\% and estimates any additional noise.
\begin{verbatim}
    >>> from nubo.models import GaussianProcess, fit_gp
    >>> from gpytorch.likelihoods import FixedNoiseGaussianLikelihood
    >>> 
    >>> 
    >>> noise = torch.ones(x_train.size(0)) * 0.025
    >>> likelihood = FixedNoiseGaussianLikelihood(noise = noise, 
    ...                                           learn_additional_noise = True)
    >>> gp = GaussianProcess(x_train, y_train, likelihood = likelihood)
    >>> fit_gp(x_train, y_train, gp = gp, likelihood = likelihood)
\end{verbatim}

\subsection{Bayesian optimisation} \label{sec:nubo-bo}

Before describing the individual optimisation options in detail, we want to 
illustrate NUBO's user experience, that is, its easy-to-read and write 
syntax, flexibility, and modularity, on a simple Bayesian optimisation step that 
can be further divided into four substeps.

First, we define the input space. Here, we want to optimise a six-dimensional 
objective function that is bounded by the hyper-rectangle $[0, 1]^6$ specified 
as \code{bounds}, a $2 \times 6$ \code{torch.Tensor}, where the first row 
provides the lower bounds and the second row the upper bounds for all six input 
dimensions. Second, we load the training inputs \code{x\_train} and the training 
outputs \code{y\_train}. This training data can be selected manually or generated 
by using a space-filling design, such as Latin hypercube sampling introduced in 
Section~\ref{sec:nubo-utils}. Third, we define and train the Gaussian process 
implemented in NUBO as discussed in Section~\ref{sec:nubo-gp}, or set up a 
custom Gaussian process with GPyTorch. Fourth, we specify an acquisition 
function that takes the fitted Gaussian process as an argument and chooses an 
optimisation method. In this case, we use the upper confidence bound introduced 
in Equation~\ref{eq:acq-ucb} and optimise it with the L-BFGS-B algorithm 
\citep{Zhu1997} using the \code{single} function.
\begin{verbatim}
    >>> import torch
    >>> from nubo.acquisition import UpperConfidenceBound
    >>> from nubo.models import GaussianProcess, fit_gp
    >>> from nubo.optimisation import single
    >>> from gpytorch.likelihoods import GaussianLikelihood
    >>> 
    >>> 
    >>> bounds = torch.tensor([[0., 0., 0., 0., 0., 0.],
    ...                        [1., 1., 1., 1., 1., 1.]])
    >>> 
    >>> x_train = # load inputs as torch.Tensor
    >>> y_train = # load outputs as torch.Tensor
    >>>    
    >>> likelihood = GaussianLikelihood()
    >>> gp = GaussianProcess(x_train, y_train, likelihood = likelihood)
    >>> fit_gp(x_train, y_train, gp = gp, likelihood = likelihood)
    >>> 
    >>> acq = UpperConfidenceBound(gp = gp, beta = 4)
    >>> x_new, _ = single(func = acq, method = "L-BFGS-B", bounds = bounds)
\end{verbatim}
NUBO is very flexible and allows the user to swap out individual elements 
for other options. For example, we can substitute the \code{UpperConfidenceBound} 
acquisition function or the \code{single} optimisation strategy without changing 
any of the other lines of code. This makes it easy and fast to tailor Bayesian
optimisation to specific problems.

\begin{figure}[t!]
\label{fig:flowchart}
\centering
\includegraphics[width=0.5\linewidth]{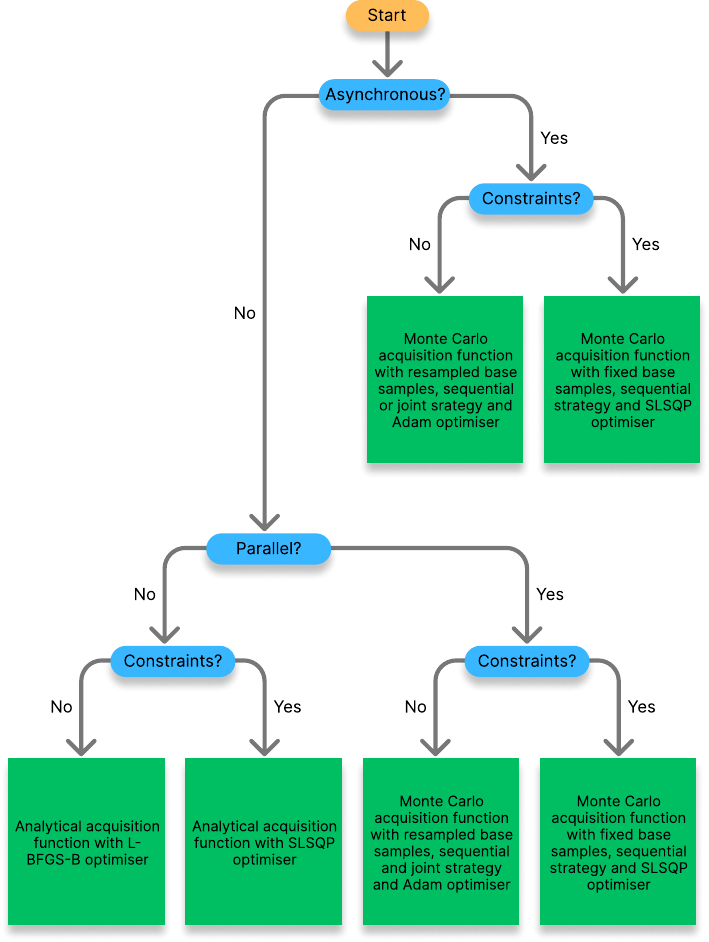}
\caption{NUBO flowchart. Overview of the recommended algorithms for specific 
problems. Start in yellow, decisions in blue, and recommended algorithm in green.}
\end{figure}

The remainder of this section introduces NUBO's optimisation strategies. 
Figure~\ref{fig:flowchart} shows a flowchart that helps users decide on the right 
acquisition function and optimiser for their specific problem.

\subsubsection{Sequential single-point optimisation}

In NUBO we differentiate between two optimisation strategies: single-point 
and multi-point optimisation. When using the single-point strategy via the 
\code{single} function, NUBO uses the analytical acquisition functions 
discussed in Section~\ref{sec:bo-acq} to find the next point to be evaluated by 
the objective function. The corresponding observation must be gathered before the 
next iteration of the optimisation loop can begin.

The code below shows how the analytical expected improvement (EI) and the 
analytical upper confidence bound (UCB) can be specified with NUBO. The 
former takes the best training output to date as the argument \code{y\_best}, 
while the latter accepts the trade-off hyper-parameter $\beta$ as the 
\code{beta} argument. For bounded optimisation problems with analytical 
acquisition functions, the optimisation method of the \code{single} function 
should be set to \code{method = "L-BFGS-B"} and the arguments \code{num\_starts} 
(default 10) and \code{num\_samples} (default 100) can be set to enable 
multi-start optimisation, where the selected optimisation algorithm is run 
multiple times and each start is initialised at the best points from a large 
number of points sampled from a Latin hypercube. This reduces the risk of getting 
stuck in a local optimum. Section~\ref{sec:nubo-utils} introduces Latin hypercube 
sampling in more detail. The \code{single} function only returns the best start 
and its acquisition value. Sequential single-point optimisation can be paired 
with constrained and mixed optimisation, which are all detailed in this section.
\begin{verbatim}
    >>> from nubo.acquisition import ExpectedImprovement, UpperConfidenceBound
    >>> from nubo.optimisation import single
    >>> 
    >>> 
    >>> acq = ExpectedImprovement(gp = gp, y_best = torch.max(y_train))
    >>> acq = UpperConfidenceBound(gp = gp, beta = 4)
    >>> x_new, _ = single(func = acq, method = "L-BFGS-B", bounds = bounds,
    ...                   num_starts = 5, num_samples = 50)
\end{verbatim}

\subsubsection{Parallel multi-point optimisation}

The second optimisation strategy is multi-point optimisation. This strategy uses 
the Monte Carlo acquisition functions outlined in Section~\ref{sec:bo-acq} to 
find multiple points, also called batches, in each iteration of the Bayesian 
optimisation loop. This strategy is particularly beneficial for objective 
functions that support parallel evaluations as points can be queried 
simultaneously, speeding up the optimisation process.

NUBO uses the Monte Carlo versions of expected improvement 
\code{MCExpectedImprovement} and upper confidence bound 
\code{MCUpperConfidenceBound} in unison with either the \code{multi\_joint} or 
\code{multi\_sequential} function to compute batches. The two different options 
for the multi-point optimisation strategy are discussed in 
Section~\ref{sec:bo-acq}. In addition to the arguments of the analytical 
acquisition functions, both Monte Carlo acquisition functions accept the number 
of Monte Carlo samples to be used to approximate the acquisition function as the 
\code{samples} argument (default 512). For the optimisation functions, the 
number of points to be computed can be passed to the \code{batch\_size} argument, 
while the \code{method} should be set to \code{"Adam"} to enable stochastic 
optimisation via the Adam algorithm \citep{Kingma2014}. The Adam algorithm can be 
fine-tuned by setting the learning rate \code{lr} (default 0.1) and the number of 
optimisation steps \code{steps} (default 100). Parallel multi-point optimisation 
can be paired with asynchronous, constrained, and mixed optimisation, which are 
all detailed in this section.
\begin{verbatim}
    >>> from nubo.acquisition import MCExpectedImprovement,
    ...                              MCUpperConfidenceBound
    >>> from nubo.optimisation import multi_joint, multi_sequential
    >>> 
    >>> 
    >>> acq = MCExpectedImprovement(gp = gp, y_best = torch.max(y_train),
    ...                             samples = 256)
    >>> acq = MCUpperConfidenceBound(gp = gp, beta = 4, samples = 256)
    >>> x_new, _ = multi_joint(func = acq, method = "Adam", lr = 0.1,
    ...                        steps = 100, batch_size = 4, bounds = bounds)
    >>> x_new, _ = multi_sequential(func = acq, method = "Adam", lr = 0.1,
    ...                        steps = 100, batch_size = 4, bounds = bounds)
\end{verbatim}
To enable the use of deterministic optimisers, such as L-BFGS-B \citep{Zhu1997} 
and SLSQP \citep{Kraft1994}, the base samples used to compute the Monte Carlo 
samples can be fixed by setting \code{fix\_base\_samples = True} (default 
\code{False}).
\begin{verbatim}
    >>> from nubo.acquisition import MCUpperConfidenceBound
    >>> from nubo.optimisation import multi_joint, multi_sequential
    >>> 
    >>> 
    >>> acq = MCUpperConfidenceBound(gp=gp, beta=4, fix_base_samples=True)
    >>> x_new, _ = multi_joint(func=acq, method="L-BFGS-B",
    ...                        batch_size=4, bounds=bounds)
    >>>
    >>> acq = MCUpperConfidenceBound(gp=gp, beta=4, fix_base_samples=True)
    >>> x_new, _ = multi_sequential(func=acq, method="L-BFGS-B",
    ...                        batch_size=4, bounds=bounds)
\end{verbatim}

\subsubsection{Asynchronous optimisation}

NUBO supports asynchronous optimisation, that is, the continuation of the 
optimisation loop while some points are being evaluated by the objective function. 
In this case, the Monte Carlo acquisition functions \code{MCExpectedImprovement} 
or \code{MCUpperConfidenceBound} are used as outlined in Section~\ref{sec:bo-acq}.
The code snippet below assumes that the two points \code{x\_pend} are currently 
in the evaluation process. To continue the optimisation, these points can be fed 
into the acquisition function by setting \code{x\_pending = x\_pend} and NUBO 
will take them into account for the subsequent iteration.
\begin{verbatim}
    >>> import torch
    >>> from nubo.acquisition import MCUpperConfidenceBound
    >>> from nubo.optimisation import multi_joint, multi_sequential
    >>> 
    >>> 
    >>> x_pend = torch.tensor([[0.2, 0.9, 0.8, 0.4, 0.4, 0.1],
    ...                        [0.1, 0.3, 0.7, 0.2, 0.1, 0.2]])
    >>> acq = MCUpperConfidenceBound(gp = gp, beta = 4, x_pending = x_pend)
    >>> x_new, _ = multi_joint(func = acq, method = "Adam",
    ...                        batch_size = 4, bounds = bounds)
    >>> x_new, _ = multi_sequential(func = acq, method = "Adam",
    ...                        batch_size = 4, bounds = bounds)
\end{verbatim}
While Monte Carlo acquisition functions are approximations of the analytical 
functions, they are mainly used for computing multiple points, where analytical 
functions are generally intractable. The Monte Carlo approach can also be used 
for single-point asynchronous optimisation by setting \code{batch\_size = 1}.

\subsubsection{Constrained optimisation}

The simple maximisation problem in Equation~\ref{eq:max-problem} can be extended 
by including one or more input constraints
\begin{equation} \label{eq:constrained-problem}
\begin{aligned}
& \boldsymbol x^* = \arg \max_{\boldsymbol x \in \mathcal{X}} f(\boldsymbol x), \\
\text{subject to} \quad & g_i(\boldsymbol x) = 0   && 
    \forall i = 1, \dots, I && \text{[Equality constraint]} \\
& h_j(\boldsymbol x) \ge 0 && \forall j = 1, \dots, J && 
    \text{[Inequality constraint].}
\end{aligned}
\end{equation}
In these instances, NUBO allows constrained Bayesian optimisation by using 
the SLSQP algorithm to optimise the acquisition function. Implementing this 
method requires the additional step of specifying the constraints \code{cons} as 
a dictionary for one constraint or a list of dictionaries for multiple 
constraints. Each constraint requires two entries. The first is \code{"type"} and 
can either be set to \code{"ineq"} for inequality constraints or \code{"eq"} for 
equality constraints. The second is \code{"fun"}, which takes a function 
representing the constraint. The optimiser only selects points for which the 
constraint functions are greater than or equal to zero for inequality constraints 
and exactly zero for equality constraints. The code snippet below specifies two 
constraints: The first is an inequality constraint that requires the first two 
input dimensions to be smaller than or equal to 0.5. The second is an equality 
constraint that requires dimensions four, five, and six to sum to 1.2442. 
\begin{verbatim}
    >>> import torch
    >>> 
    >>> 
    >>> bounds = torch.tensor([[0., 0., 0., 0., 0., 0.], 
    ...                        [1., 1., 1., 1., 1., 1.]])
    >>> cons = [{"type": "ineq", "fun": lambda x: 0.5 - x[0] - x[1]}, 
    ...         {"type": "eq", "fun": lambda x: 1.2442 - x[3] - x[4] - x[5]}]
\end{verbatim}
After setting up the input space using the bounds and constraints, the Bayesian 
optimisation loop is similar to before. We need to set the \code{method} argument 
of the optimisation function to \code{"SLSQP"} and provide the function with the 
constraints \code{cons}.
\begin{verbatim}
    >>> from nubo.acquisition import UpperConfidenceBound
    >>> from nubo.optimisation import single
    >>> 
    >>> 
    >>> acq = UpperConfidenceBound(gp = gp, beta = 4)
    >>> x_new, _ = single(func = acq, method = "SLSQP",
    ...                   bounds = bounds, constraints = cons)
\end{verbatim}
Constrained Bayesian optimisation can be used with analytical and Monte Carlo 
acquisition functions as well as single-point, multi-point, asynchronous, and 
mixed optimisation, all of which are detailed in this section.

\subsubsection{Mixed optimisation}

Bayesian optimisation is predominantly focused on problems with continuous input 
parameters since the Gaussian process models all input dimensions as continuous 
variables. However, NUBO supports the optimisation of mixed input parameter 
spaces via a workaround. To do this, NUBO first computes all possible 
combinations of the discrete parameters. Then, it maximises the acquisition 
function for all continuous parameters while holding one combination of the 
discrete parameters fixed. Once the acquisition function is maximised for each of 
the possible discrete combinations, the best overall solution is returned. Note 
that this can be very time-consuming for many discrete dimensions or discrete 
values.

To implement mixed optimisation in NUBO, bounds are specified as before, 
but the discrete dimensions are additionally defined in a dictionary where the 
keys are the dimensions (starting from zero) and the values are a list of all 
possible values for the discrete inputs. The code below specifies dimensions one 
and five as \code{disc}.
\begin{verbatim}
    >>> import torch
    >>> 
    >>> 
    >>> bounds = torch.tensor([[0., 0., 0., 0., 0., 0.],
    ...                        [1., 1., 1., 1., 1., 1.]])
    >>> disc = {0: [0.2, 0.4, 0.6, 0.8],
    ...         4: [0.3, 0.6, 0.9]}
\end{verbatim}
After setting up the input space specified by the bounds and discrete values, 
the Bayesian optimisation loop is similar to before. We only need to provide the 
function with the dictionary specifying the discrete dimensions 
\code{discrete=disc}.
\begin{verbatim}
    >>> from nubo.acquisition import UpperConfidenceBound
    >>> from nubo.optimisation import single
    >>> 
    >>> 
    >>> acq = UpperConfidenceBound(gp = gp, beta = 4)
    >>> x_new, _ = single(func = acq, method = "L-BFGS-B",
    ...                   bounds = bounds, discrete = disc)
\end{verbatim}
Mixed Bayesian optimisation can be used in unison with analytical and Monte Carlo 
acquisition functions as well as single-point, multi-point, asynchronous, and 
constrained optimisation, all of which are detailed in this section.

\subsection{Test functions and utilities} \label{sec:nubo-utils}

\begin{figure}[t!] \label{fig:lhs}
    \centering
    \includegraphics[width=0.7\linewidth]{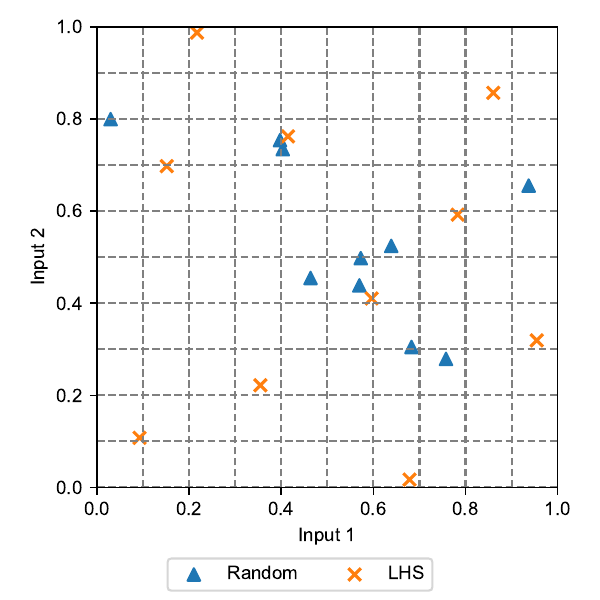}
    \caption{Latin hypercube sampling compared to random sampling.}
\end{figure}

NUBO provides a selection of test functions and utilities to make 
implementing and testing Bayesian optimisation algorithms more convenient. The 
ten test functions were selected from the virtual library of \cite{Surjanovic} 
and represent a variety of challenges, such as bowl-shaped, plate-shaped, 
valley-shaped, uni-modal, and multi-modal functions. The functions can be 
imported from the \code{nubo.test\_functions} module and instantiated by providing 
the number of dimensions (except for the Hartmann function that comes in 3D and 
6D versions), the standard deviation of any noise that should be added, and 
whether the function should be minimised or maximised. These functions are 
equipped with the following attributes: the number of dimensions~\code{dims}, the 
bounds~\code{bounds}, and the inputs and outputs of the global 
optimum~\code{optimum}.
\begin{verbatim}
    >>> from nubo.test_functions import Ackley, Hartmann6D
    >>> 
    >>> 
    >>> func = Ackley(dims = 5, noise_std = 0.1, minimise = False)
    >>> func = Hartmann6D(minimise = False)
    >>> dims = func.dims
    >>> bounds = func.bounds
\end{verbatim}
The \code{gen\_inputs} function from the \code{nubo.utils} module allows us to 
generate input data that covers the input space efficiently by sampling a larger 
number of random Latin hypercube designs \citep{McKay2000} and returning the 
design with the largest minimal distance between all points. Figure~\ref{fig:lhs} 
compares Latin hypercube sampling to random sampling for two input dimensions. 
While many random points are in close proximity to each other, points from the 
Latin hypercube design cover the whole input space effectively by only placing 
one point in each row and column. The exact position of the point within the 
selected square is random. The code snippet below generates five points for each 
input dimension of the Hartmann function initiated above and uses \code{func} to 
compute the corresponding outputs.
\begin{verbatim}
    >>> from nubo.utils import gen_inputs
    >>> 
    >>> 
    >>> x_train = gen_inputs(num_points = dims * 5,
    ...                      num_dims = dims,
    ...                      bounds = bounds)
    >>> y_train = func(x_train)
\end{verbatim}

Finally, we discuss three convenience functions that can be used for data 
transformation. \code{normalise} and \code{unnormalise} can be used to scale 
input data to the unit cube $[0, 1]^{d}$ and back to its original domain by 
providing the bounds of the input space. Furthermore, the outputs can be centred 
at zero with a standard deviation of one with the \code{standardise} function.
\begin{verbatim}
    >>> from nubo.utils import standardise, normalise, unnormalise
    >>> 
    >>> 
    >>> x_norm = normalise(x_train, bounds = bounds)
    >>> x_train = unnormalise(x_norm, bounds = bounds)
    >>> y_stand = standardise(y_train)
\end{verbatim}

%% -- Case study ---------------------------------------------------------------

\section{Case study} \label{sec:case-study}

We present the general workflow for optimising an expensive-to-evaluate black-box 
function with NUBO by providing a detailed case study in which a test 
function with six input dimensions is optimised. This case study demonstrates how 
the user can specify the parameter input space, generate initial training data, 
and define and run the Bayesian optimisation loop.

\begin{figure}[t!] 
\label{fig:case-study}
\centering
\includegraphics[width=0.7\linewidth]{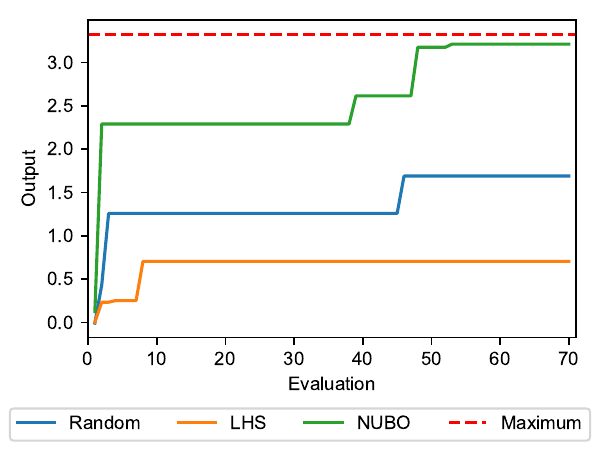}
\caption{Results of the Bayesian optimisation algorithm implemented with 
NUBO as defined in this case study compared to random sampling and Latin 
hypercube sampling.}
\end{figure}

So that the case study is reproducible, we set the seed for the pseudo-number 
generator within Torch to 123. We set some format options for the 
\code{print} function such that values are rounded to the fourth decimal place 
and are not formatted in scientific notation to increase readability.
\begin{verbatim}
    >>> import torch
    >>>  
    >>> 
    >>> torch.manual_seed(123)
    >>> torch.set_printoptions(precision = 4, sci_mode = False)
\end{verbatim} 
A typical objective function optimised with Bayesian optimisation is expensive to 
evaluate and thus not feasible to use in a case study that aims to illustrate how 
NUBO can be applied. Hence, we will use one of the synthetic test functions 
provided by NUBO as a surrogate expensive-to-evaluate black-box function. 
We use the six-dimensional Hartmann function that possesses multiple local and 
one global minimum. Its input space is bounded by the hyper-rectangle $[0, 1]^6$. 
Observational noise, such as measurement error, is represented by adding a small 
amount of random Gaussian noise to the function output by setting 
\code{noise\_std = 0.1}. \code{minimise} is set to \code{False} to transform the 
minimisation into a maximisation problem as required for Bayesian optimisation 
with NUBO.
\begin{verbatim}
    >>> from nubo.test_functions import Hartmann6D
    >>> 
    >>> 
    >>> black_box = Hartmann6D(noise_std = 0.1, minimise = False)
\end{verbatim}
With our objective function specified, we can focus on defining the input space. 
We know that our objective function has six inputs that are all bounded by 
$[0, 1]$. As introduced in Section~\ref{sec:nubo-bo}, the bounds are defined as 
a $2 \times d$ \code{torch.tensor}, where the first row specifies the lower 
bounds and the second row specifies the upper bounds. This case study also 
highlights the mixed parameter optimisation capabilities of NUBO (see 
Section~\ref{sec:nubo-bo}) by assuming that the first input is a discrete 
parameter restricted to 0.2, 0.4, 0.6, and 0.8. We can implement this by 
specifying a dictionary, where the key is the input dimension and the value is a 
list of possible values the input can take, that is 
\code{\{0: [0.0, 0.1, 0.2, 0.3, 0.4, 0.5, 0.6, 0.7, 0.8, 0.9, 1.0]\}}. Note that 
indexing starts at zero in Python.
\begin{verbatim}
    >>> dims = 6
    >>> bounds = torch.tensor([[0., 0., 0., 0., 0., 0.],
    ...                        [1., 1., 1., 1., 1., 1.]])
    >>> discrete = {0: [0.0, 0.1, 0.2, 0.3, 0.4, 0.5,
    ...                 0.6, 0.7, 0.8, 0.9, 1.0]}
\end{verbatim}
The Bayesian optimisation loop requires initial training data. This is important 
to train the Gaussian process that tries to emulate the objective function. This 
case study uses the \code{gen\_inputs} function introduced in 
Section~\ref{sec:nubo-utils} to generate 30 initial data points from a Latin 
hypercube design. We round the first input dimension to fit the discrete values 
specified above as Latin hypercube designs return continuous values. These points 
are evaluated by the objective function to complete our training data pairs 
consisting of input parameters \code{x\_train} and observations \code{y\_train}.
\begin{verbatim}
    >>> from nubo.utils import gen_inputs
    >>> 
    >>> 
    >>> x_train = gen_inputs(num_points = dims * 5,
    ...                      num_dims = dims,
    ...                      bounds = bounds)
    >>> x_train[:, 0] = torch.round(x_train[:, 0], decimals = 1)
    >>> y_train = black_box(x_train)
\end{verbatim}
Next, we specify the Bayesian optimisation algorithm we plan to use in our 
optimisation loop. We define the \code{bo} function that takes our training pairs 
\code{(x\_train, y\_train)} and returns the next candidate point \code{x\_new} which 
is evaluated by the objective function in four steps. First, we set up our 
surrogate model as the Gaussian process provided by NUBO with a Gaussian 
likelihood as discussed in Section~\ref{sec:nubo-gp}. Second, we train the 
Gaussian process \code{gp} with our training data by maximising the likelihood 
with the Adam algorithm \citep{Kingma2014} via the \code{fit\_gp} function. Here, 
we set a custom learning rate \code{lr} and the number of optimisation steps 
\code{steps}. Third, we define an acquisition function that will guide our 
optimisation. As we assume that our objective function allows parallel function 
evaluations, we decide to compute multi-point batches at each iteration and choose 
a Monte Carlo acquisition function, in this case \code{MCUpperConfidenceBound}. 
The acquisition function \code{acq} is instantiated by providing it with the 
fitted Gaussian process \code{gp}, a value for the trade-off hyper-parameter 
\code{beta}, and the number of Monte Carlo samples used to approximate the 
acquisition function. For further details, refer to Section~\ref{sec:bo-acq}. 
Fourth, we maximise the acquisition function \code{acq} with the 
\code{multi\_sequential} function that uses the sequential strategy for computing 
multiple candidate points. We decide to compute four candidate points at each 
iteration by setting \code{batch\_size=4} and providing the previously specified 
bounds and discrete values. The Adam optimiser is used as Monte Carlo acquisition 
functions require a stochastic optimiser due to their inherent randomness 
introduced by drawing the Monte Carlo samples. The optimiser is initialised at 
two different initial points chosen as the two points with the highest 
acquisition value out of 100 potential points sampled from a Latin hypercube 
design. We chose two initialisations to keep the computational overhead within 
the replication script low. In practice, a higher number of initialisations might 
be beneficial.
\begin{verbatim}
    >>> from nubo.acquisition import MCUpperConfidenceBound
    >>> from nubo.models import GaussianProcess, fit_gp
    >>> from nubo.optimisation import multi_sequential
    >>> from gpytorch.likelihoods import GaussianLikelihood
    >>> 
    >>> 
    >>> def bo(x_train, y_train):
    >>> 
    >>>     likelihood = GaussianLikelihood()
    >>>     gp = GaussianProcess(x_train, y_train, likelihood = likelihood)
    >>>
    >>>     fit_gp(x_train, y_train, gp = gp, likelihood = likelihood,
    ...            lr = 0.1, steps = 200)
    >>> 
    >>>     acq = MCUpperConfidenceBound(gp = gp, beta = 4, samples = 128)
    >>> 
    >>>     x_new, _ = multi_sequential(func = acq,
    ...                                 method = "Adam",
    ...                                 batch_size = 4,
    ...                                 bounds = bounds,
    ...                                 discrete = discrete,
    ...                                 lr = 0.1,
    ...                                 steps = 200,
    ...                                 num_starts = 2,
    ...                                 num_samples = 100)
    >>>     
    >>>     return x_new
\end{verbatim}
Finally, we specify the entire optimisation loop, that is, a simple 
\code{for}-loop that computes the next batch of candidate points using the defined 
Bayesian optimisation algorithm \code{bo}, evaluates the candidate points by the 
objective function \code{black\_box}, and adds the new data pairs 
\code{(x\_new, y\_new)} to the training data. We let the optimisation loop run for 
ten iterations and print all evaluations, where the first six columns are the 
inputs and the final column is the output from the objective function. The first 
30 rows give the initial training data generated by the Latin hypercube design, 
while the last 40 rows were chosen by the Bayesian optimisation algorithm. The 
results clearly show that NUBO improves upon the initial space-filling 
design and produces points which are consistent with the bounds and discrete 
values that specify the parameter input space.
\begin{verbatim}
        >>> iters = 10
        >>> 
        >>> for iter in range(iters):
        >>> 
        >>>     x_new = bo(x_train, y_train)
        >>> 
        >>>     y_new = black_box(x_new)
        >>> 
        >>>     x_train = torch.vstack((x_train, x_new))
        >>>     y_train = torch.hstack((y_train, y_new))
        >>> 
        >>> print(torch.hstack([x_train, y_train.reshape(-1, 1)]))
        
        tensor([[0.2000, 0.6523, 0.1574, 0.7822, 0.3039, 0.8603,  0.1251],
                [0.5000, 0.9127, 0.8746, 0.4787, 0.6523, 0.1249,  2.2907],
                [0.4000, 0.5638, 0.0459, 0.6200, 0.7056, 0.2929,  0.6744],
                [0.2000, 0.3003, 0.2290, 0.8110, 0.9529, 0.2384,  0.0442],
                [0.1000, 0.7809, 0.5374, 0.1381, 0.5655, 0.5679,  0.6123],
                ...
                [0.5000, 1.0000, 1.0000, 0.5839, 0.0000, 0.3425,  0.7659],
                [0.4000, 0.8899, 1.0000, 0.5622, 0.0000, 0.0421,  3.1330],
                [0.4000, 0.8026, 0.0000, 0.5356, 0.0000, 0.0000,  2.7828],
                [0.2000, 0.0556, 1.0000, 1.0000, 0.0000, 0.6695,  0.1134],
                [0.4000, 0.7124, 1.0000, 0.6842, 0.0000, 0.0000,  2.3028]],
                dtype=torch.float64)
\end{verbatim}
NUBO explores the parameter space efficiently by switching between exploring 
areas with high uncertainty and areas with high predictions. This means that the 
algorithm does not monotonically converge to a single solution as conventional 
optimisation algorithms would. Thus, the approximate solution to an objective 
function is the best value found during optimisation. In this case study, the 
approximate solution, i.e., the solution with the highest output (last column in 
the Python output above), was found at iteration 53, and the inputs 
and outputs are printed below.
\begin{verbatim}
        >>> best_iter = int(torch.argmax(y_train))
        >>> 
        >>> print("Approximate solution")
        >>> print("--------------------")
        >>> print(f"Evaluation: {best_iter + 1}")
        >>> print(f"Inputs: {x_train[best_iter]}")
        >>> print(f"Output: {y_train[best_iter]:.4f}")

            Approximate solution
        --------------------
        Evaluation: 53
        Inputs: tensor([0.4000, 0.9136, 1.0000, 0.5669, 0.0000, 0.0802],
                       dtype=torch.float64)
        Output: 3.2133
\end{verbatim}
We compare the results provided by NUBO with the results from random 
sampling and using a space-filling design, in this case, Latin hypercube sampling 
(LHS). The code below generates results for the full budget of 70 evaluations for 
both sampling methods and plots the results against each other, with the number 
of evaluations on the x-axis and the accumulative best output for each method on 
the y-axis. Figure~\ref{fig:case-study} shows that NUBO (green line) 
provides a better solution than either alternative approach and is very close to 
the true maximum of 3.32237. NUBO succeeds in accurately approximating the 
true optimum.

\begin{verbatim}
    >>> import matplotlib.pyplot as plt
    >>> import numpy as np
    >>> 
    >>> 
    >>> torch.manual_seed(123)
    >>> random = black_box(torch.rand((70, dims)))
    >>> lhs = black_box(gen_inputs(num_points = 70,
    ...                            num_dims = dims,
    ...                            bounds = bounds))
    >>> 
    >>> plt.plot(range(1, 71), np.maximum.accumulate(random), label = "Random")
    >>> plt.plot(range(1, 71), np.maximum.accumulate(lhs), label = "LHS")
    >>> plt.plot(range(1, 71), np.maximum.accumulate(y_train), label = "NUBO")
    >>> plt.hlines(3.32237, 0, 71, colors = "red", linestyles = "dashed",
    ...            label = "Maximum")
    >>> plt.title("Comparison against random designs")
    >>> plt.xlabel("Evaluations")
    >>> plt.ylabel("Output")
    >>> plt.legend(loc = 'lower center', ncol = 4, bbox_to_anchor = (0.5, -0.275))
    >>> plt.xlim(0, 71)
    >>> plt.tight_layout()
\end{verbatim}
%

%% -- Summary/conclusions/discussion -------------------------------------------

\section{Conclusion} \label{sec:conclusion}

This article introduces NUBO, a Python package for Bayesian 
optimisation to optimise expensive-to-evaluate black-box functions, for example, 
computer simulators and physical experiments. The main objective of NUBO is 
to make Bayesian optimisation accessible to researchers from all disciplines by 
providing a transparent and user-friendly implementation.

NUBO includes five sub-modules that implement Gaussian processes, 
acquisition functions, optimisers, test functions, and utilities. These modules 
provide all necessities for sequential single-point, parallel multi-point, and 
asynchronous optimisation of expensive-to-evaluate black-box functions for 
bounded, constrained, and/or mixed (discrete and continuous) input parameter 
spaces. We have introduced and explained each of these functionalities with 
individual code snippets and illustrated NUBO's general workflow using a 
detailed case study that takes a hypothetical six-dimensional 
expensive-to-evaluate black-box function and approximates its global optimum with 
a parallel multi-point Bayesian optimisation algorithm. 

A brief comparison with other Python packages for Bayesian optimisation 
showed that NUBO has competitive performance while providing a transparent 
and simple implementation. This makes NUBO a good candidate for the 
optimisation of expensive black-box functions when transparency is vital.

In the future, we plan to extend NUBO to include optimisation strategies 
for multi-fidelity, multi-objective, and high-dimensional problems.

\section*{Computational details}

The results in this paper were obtained using Python~3.11.2 with the 
following packages: NUBO~1.0.3, Torch~2.0.0, GPyTorch~1.10 , 
SciPy~1.10.1, NumPy~1.24.2, and Matplotlib~3.9.0 
\citep{mtatplotlib}. For the package comparison bayes\_opt~1.4.3, 
BoTorch~0.8.4, pyGPGO~0.5.0 and SMAC3~2.0.0 were used.  
Python itself is available from the Python website at 
\url{https://www.python.org/} and all packages used are available from the
Python Package Index (PyPI) at \url{https://www.pypi.org/}.

\section*{Acknowledgments}

The work has been supported by the Engineering and Physical Sciences Research 
Council (EPSRC) under grant number EP/T020946/1 and the EPSRC Centre for Doctoral 
Training in Cloud Computing for Big Data under grant number EP/L015358/1. 

%Bibliography
\bibliographystyle{unsrt}  
\bibliography{references}

\end{document}